\documentclass{bmvc2k}

\usepackage{booktabs}
\usepackage{bbm}
\usepackage{amsmath}
\DeclareMathOperator*{\argmax}{arg\,max}

\usepackage{algpseudocode}
\usepackage{algorithm}
\usepackage{multirow}
\usepackage{wrapfig, graphicx}

\title{CLAD: A Contrastive Learning based Approach for Background Debiasing}

\addauthor{Ke Wang}{k.wang@epfl.ch}{}
\addauthor{Harshitha Machiraju$^\star$}{harshitha.machiraju@epfl.ch}{1,2}
\addauthor{Oh-Hyeon Choung$^\star$}{ohhyeon.choung@gmail.com}{1}
\addauthor{Michael H. Herzog}{michael.herzog@epfl.ch}{1}
\addauthor{Pascal Frossard}{pascal.frossard@epfl.ch}{2}

\addinstitution{
 Laboratory of Psychophysics (LPSY)\\
 Ecole Polytechnique Fédérale de Lausanne (EPFL), \\
 Switzerland
}
\addinstitution{
 Signal Processing Laboratory 4 (LTS4)\\
 Ecole Polytechnique Fédérale de Lausanne (EPFL), \\
 Switzerland
}

\runninghead{Wang et al }{CLAD}

\begin{document}

\maketitle

\begin{abstract}
Convolutional neural networks (CNNs) have achieved superhuman performance in multiple vision tasks, especially image classification. However, unlike humans, CNNs leverage spurious features, such as background information to make decisions. This tendency creates different problems in terms of robustness or weak generalization performance. Through our work, we introduce a contrastive learning-based approach (CLAD) to mitigate the background bias in CNNs. CLAD encourages semantic focus on object foregrounds and penalizes learning features from irrelavant backgrounds. Our method also introduces an efficient way of sampling negative samples. We achieve state-of-the-art results on the Background Challenge dataset, outperforming the previous benchmark with a margin of 4.1\%. Our paper shows how CLAD serves as a proof of concept for debiasing of spurious features, such as background and texture (in supplementary material). \footnote{ Code: \url{https://github.com/wangke97/CLAD.}} 
\end{abstract}

\section{Introduction}
\label{sec:intro}
CNNs have achieved superhuman performance on various computer vision tasks such as segmentation \cite{seg_sota}, classification \cite{voulodimos2018deep,he2016deep,vit_mlp}, object detection\cite{detect_sota}, etc. However, it has been observed that CNNs have a different understanding of images in contrast to humans \cite{geirhos2017comparing}. Specifically, in the case of classification, it has been observed that CNNs can be biased towards the background information instead of the foreground object \cite{zhu2016object, barbu2019objectnet, beery2018recognition, xiao2020noise, sehwag2020time}, high-frequency components \cite{freq_main,guille_freq,our_freq}, and textures rather than shapes \cite{geirhos2018imagenet,islam2021shape}. In particular, Kai et al.\cite{xiao2020noise} showed that CNNs tend to correlate class labels heavily with background information. Further, they showed that, when the foreground object is removed, CNNs still perform surprisingly well solely in the presence of the background of the image. The authors created the Background Challenge \cite{xiao2020noise} which measures models' robustness to various changes the background. They further showed that most state-of-the-art image classification models exhibit a poor generalization ability in this challenge due to large background bias.
These biases lead to an over-dependence of the model on irrelevant/spurious features. Further, such biases can be exploited to fool the classifiers by simply altering the background of the object \cite{rival_background_study} or adding different textures to the image \cite{geirhos2018imagenet}. 

To mitigate such biases, conventional data augmentation is often used, wherein the model is exposed to additional training data, in order to decorrelate the spurious features and the class label. However, to completely eliminate bias and prevent memorization (overfitting) of data, the model usually requires a very large amount of data for augmentation. Also, previous works \cite{wang2019balanced, zhao2019inherent} have shown that conventional data augmentation is insufficient to discard spurious features and remains susceptible to small changes. Therefore, an effective data augmentation method has to be applied to ensure that the best features are extracted during training while introducing minimal computational overhead.

\begin{figure*}[ht]
\centering
\includegraphics[width=0.8\textwidth]{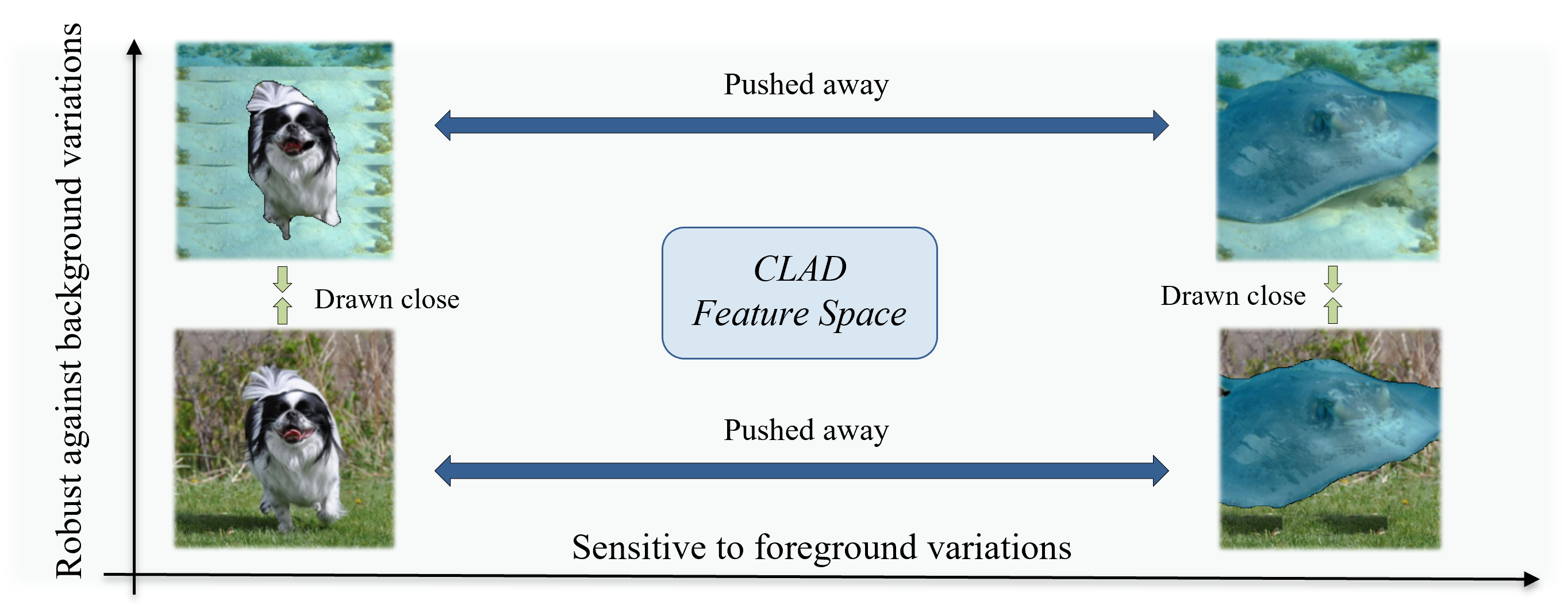}
\caption{CLAD learns feature space which is robust against background variations and sensitive to foreground features.}
\label{fig:intro} 
\end{figure*}  

Here, we propose a \textbf{C}ontrastive \textbf{L}earning based \textbf{A}pproach for background \textbf{D}ebiasing (CLAD) model, where contrastive learning (CL) is introduced to mitigate the biases more effectively. In contrastive learning, for each data point (anchor), both positive samples (sharing anchor's distribution) and negative samples (carries different information as anchor) are generated. Then, CL minimizes the distances between the anchor and positive samples and maximizes the distances between the anchor and negative samples in feature space. To this end, in the latent embedding, attributes that belong to the same distribution (relevant features, e.g., foregrounds) are aggregated together, while unwanted biases (e.g., backgrounds) are separated from the anchor. Hence, our method, CLAD, uses contrastive learning to learn a background-robust feature space, by carefully constructing the positive and negative samples. Positive samples are generated by changing the anchor's background. The negative samples, on the other hand, contain distinctive foreground information but similar background as the anchor (Fig. \ref{fig:construct samples}). Moreover, instead of generating negative samples, we introduce a novel mechanism to sample negative samples without introducing extra costs, where we sample negative pairs during the training process from generated positive samples while ensuring the sampled negative samples share similar background information as anchor. Thus, our novel method allows for both the scalability of negative samples as well as having similar background as anchor. \\
We show that, our CLAD model \textbf{outperforms} the state-of-the-art methods on the Background Challenge dataset \cite{xiao2020noise}, while it has almost no accuracy drop on original images. It samples negative samples effectively without introducing heavy computational costs. Especially, CLAD outperforms on the random-background dataset (\textsc{Mixed-Rand}) by a margin of $4.1\%$, while all the other state-of-the-art methods showed a major performance drop. We also show that CLAD can be applied to mitigate the influence of other discriminative features apart from background, like object texture (in supplementary material), while improving model's shape bias.

\section{Related Work}
\label{seg:related_work}

Feature biases are thought to happen because of the data memorization (overfitting) and are exacerbated when training the over-parameterized models \cite{Khani2021spurious}. One effective way to mitigate these problems is to augment with samples emphasizing desirable features instead of irrelevant spurious ones. In background-biased settings, Kai et al.\cite{xiao2020noise} showed that training models on images with random unrelated backgrounds for a given foreground helped reduce the background bias of the model. However, this also significantly reduced performance on the original dataset (Table \ref{table:acc_benchmark}). Further, as mentioned in the previous section, conventional data augmentation is not optimal for debiasing; hence, we rather look at contrastive learning. 
% Our proposed method uses Contrastive Learning, which capitalizes on this idea while also selecting positive and negative samples more efficiently, leading to better performance.

\textbf{Contrastive Learning (CL)} \cite{hadsell2006dimensionality}  helps learn robust feature spaces that are close across a data distribution and attributes that set apart a data distribution from another. CL has shown great promise in self-supervised regimes \cite{chen2020simple, he2020momentum, grill2020bootstrap} while recently, it has also been applied to the supervised learning domain and achieved promising results \cite{khosla2020supervised, lo2021clcc, liu2021social}. CL has been used in a self-supervised manner to help debias models \cite{taghanaki2021robust, lee2022improving, ryali2021leveraging, mo2021object}. 
In the fully supervised learning domain, previous works have shown that utilizing contrastive loss as an auxiliary loss can encourage learning more robust features with higher generalization abilities through careful contrastive pair construction \cite{lo2021clcc, liu2021social}. To the best of our knowledge, we are the first to leverage contrastive learning as an auxiliary loss to improve the model's background robustness in a fully-supervised setting.

\section{Methodology}
\label{Sec:Methodology}

% Models trained in a fully-supervised manner suffer from strong background bias, limiting their generalization ability to unseen data, due to overfitting to the spurious correlation between image backgrounds and class labels in the train set. Therefore, we present CLAD to mitigate background bias by destroying the background-label correlations in the train set and punishes extracting feature from image backgrounds. 

In this section, we go through the contrastive learning framework and then introduce our background-debiased contrastive pair sampling strategy, and finally present our overall learning framework.

\subsection{Contrastive Learning}
We use the popular InfoNCE \cite{gutmann2010noise, oord2018representation} loss as our contrastive loss term. This loss function can be viewed as an (N+1)-way cross-entropy classification loss to distinguish between one positive sample and N negative samples, and is written as:

\begin{equation}
\mathcal{L}_{con} = -log\left [\frac{e^{s(x, x^+)/\tau}}{e^{s(x, x^+)/\tau} + \sum_{i=1}^{N}e^{s(x, x_i^-)/\tau}} \right ]  
\label{eq:contrastive loss}
\end{equation}
where \(s(x_1, x_2) = (x_1 \cdot x_2) / (\left\|x_1 \right\| \left\|x_2 \right\|\)) is the cosine similarity function and $\tau$ is the temperature parameter; $x, x^+, x^-_i$ represent the feature representations for the anchor, the positive sample and the multiple negative samples, respectively. It brings positive sample pairs closer in the feature space, while it pushes the anchor apart from negative samples.

% In contrastive learning, a crucial task is the choices for positive and negative samples \cite{chen2020simple, kalantidis2020hard, tian2020makes, robinson2020contrastive}. Positive samples and negative samples are expected to be not easily distinguishable to for learning robust features and generalization. 

\subsection{Background-debiased Sampling}
\label{sample_sec}

\begin{figure*}[ht]
\centering
\includegraphics[width=1.03\textwidth]{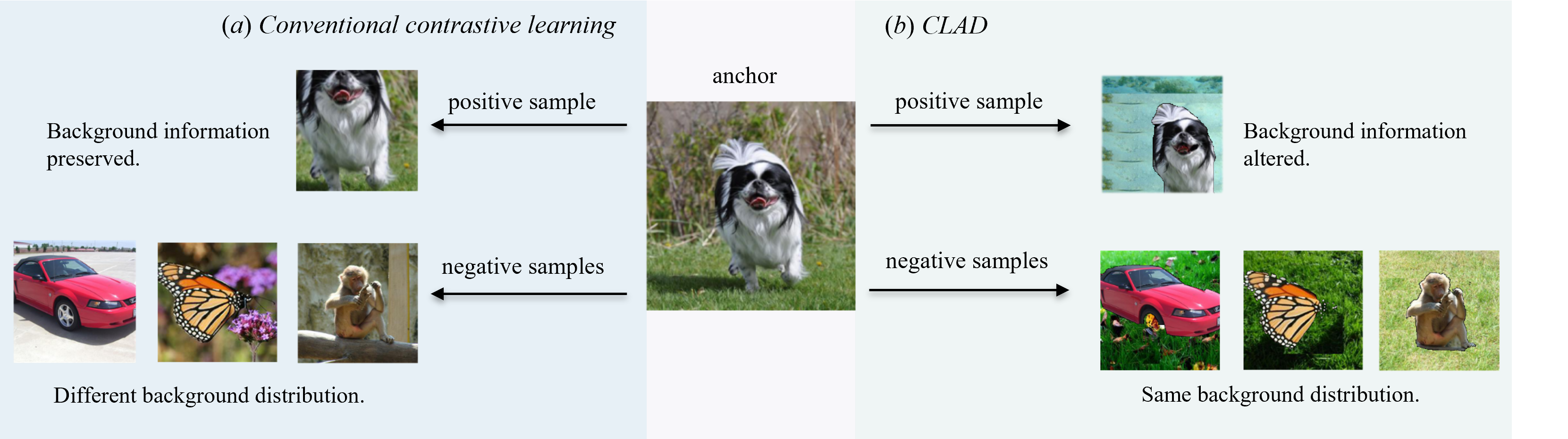}
\vspace{-1.5em}
\caption{Contrastive pair sampling strategies, (a) used in conventional contrastive learning, (b) CLAD's background-debiased sampling strategy}
\label{fig:construct samples} 
\end{figure*}

One crucial contribution of CLAD is an efficient sampling approach for contrastive pairs which are harder to discrminate from the anchor. Conventionally, in contrastive learning, positive samples are obtained by applying a combination of different data augmentations to the anchor. Negative samples, on the other hand, come from views of other images (see Fig. \ref{fig:construct samples} (a)). However, such sampling of contrastive pairs would lead to poor robustness on backgrounds due to two reasons: 1) increasing feature similarity between positive pairs would simultaneously encourage background bias due to their shared background information; 2) likewise, as negative samples carry different background information compared to the anchor, minimizing feature similarity between negative pairs would increase the model's sensitivity to background variations.

% As shown in the previous section, while contrastive learning provides a more effective strategy for data augmentation it is also computationally very expensive. In addition, most of the previous methods do not make use of the readily available label information for training the networks. 
These problems are solved in CLAD's background-debiased contrastive pair sampling approach, where background information is no longer shared between positive pairs, and negative pairs share similar background information, as shown in Fig. \ref{fig:construct samples} (b). The contrastive pairs are created as follows:

\textbf{Positive Samples}  are created by replacing the background of the anchor with a different-class background (chosen randomly). Following the method in Background Challenge dataset \cite{xiao2020noise}, we use GrabCut \cite{rother2004grabcut} to separate the foreground and background of a given anchor image (see supplementary material for details). The foreground of the anchor is then placed in a background found in another random class (other than the anchor class).
% the foreground is pre-segmented with GrabCut \cite{rother2004grabcut} while the background is generated by removing foreground and filling the missing parts with rest of the image. The positive sample is obtained by masking the background with foreground mask and pasting the foreground: $FG + {M_{FG}}*BG$.

\textbf{Negative Sample } It is crucial to have a large number of negative samples in contrastive learning \cite{chen2020simple, he2020momentum}. However, using the same method to create positive samples, i.e., replacing the foreground of the anchor image instead and keeping the background, needs to be repeated many times to create multiple negative samples. This leads to a high computational cost which linearly scales the cost per batch by the number of negative samples. To solve this issue, we introduce a negative sample dictionary. 

We define our negative sample dictionary as a dictionary with queues for each class, containing the latent representation for each negative sample. Each queue, has samples whose background belongs to the class represented by the queue. The size of each queue is the same as the number of negative samples (\textit{N}). In each batch, we use the generated positive samples to update the queue. The old samples are dequeued (deleted) when new samples are enqueued (added) to the queue following a first in, first out order. Therefore, the negative samples are reused until they get replaced in the queue.

This differs from the commonly used memory bank \cite{he2020momentum} for storing negative samples in two ways:
\vspace{-0.45em}
\begin{itemize}
    \itemsep-0.5em
    \item It only stores features for background-augmented images where the foreground and background classes are decoupled. 
    \item As a dictionary, it contains keys for \textbf{background labels} of the stored samples. Samples are stored in the queue whose key corresponds to their \textbf{background labels}.
\end{itemize}
\vspace{-0.45em}

% In each batch, we use the feature representations for the generated positive samples to update the dictionary. Each sample will enter the queue according to the labels of its background. Thus, the samples stored in each queue contain similar background information, \textit{which all comes from the distribution of key's class.}

\begin{figure*}[ht]
\centering
\includegraphics[width=\textwidth]{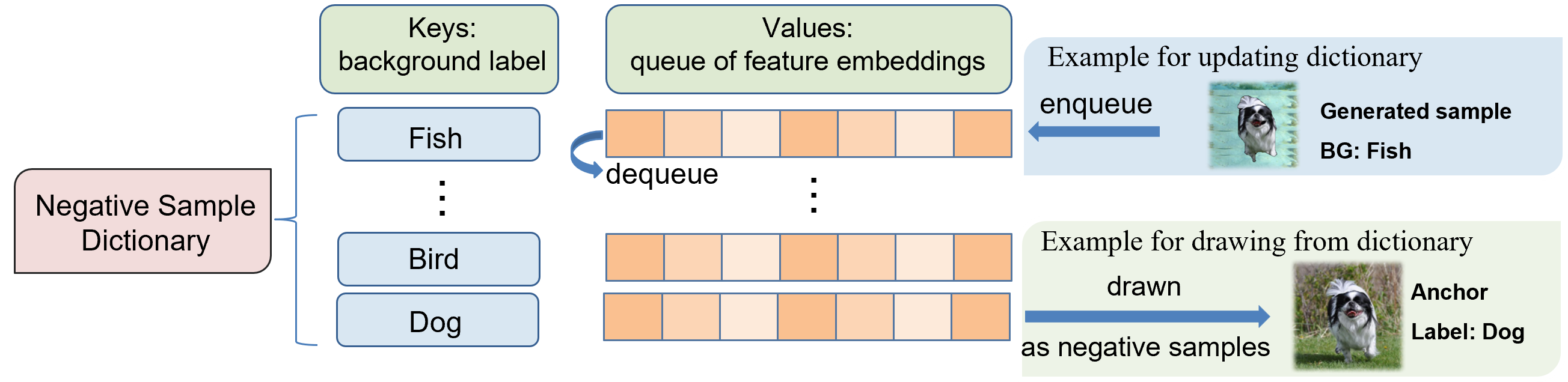}
\vspace{-1.5em}
\caption{Illustration of using dictionary to store negative sample candidates.}
\label{fig:dictionary} 
\end{figure*}

We illustrate the mechanism in Fig. \ref{fig:dictionary}. The dictionary contains the keys of the background label, and we show two examples in the Figure. In the example for updating the dictionary with generated samples, the sample has a background of \texttt{Fish} class, so it will enter the queue within the \texttt{Fish} key (the foreground label is ignored in this process). The other example shows the sampling process for negative samples from the dictionary: the anchor is an image from \texttt{Dog} class; hence we draw all samples in the queue within the \texttt{Dog} key in the dictionary.

Using the negative sample dictionary guarantees that similar background information is shared between negative pairs simultaneously. Hence, our method provides a memory-efficient way of scaling negative samples.

\subsection{Training Objective}
\label{sec:loss}

The overall loss function is composed of two terms: the conventional supervised classification loss (for learning distinguishable features) and contrastive loss (for improving background robustness). After we generate positive and negative sample pairs (as described in Sec \ref{sample_sec}), we calculate the contrastive loss using the InfoNCE loss function. 
To enforce the correct classification of the positive samples, we can optionally include a classification loss for such samples and refer to the model with this additional loss term as CLAD+. For the supervised classification loss, we use the conventional cross-entropy loss. Specifically,

the overall loss for CLAD can be written as:
\vspace{-0.5em}
\begin{equation}
\mathcal{L}_{CLAD} = \mathcal{L}_{class}(x) + \lambda * \mathcal{L}_{con}(x, x^+, x^-) 
\label{eq:loss_clad}
\end{equation}

For CLAD+, the loss is written as:
\vspace{-0.5em}
\begin{equation}
\mathcal{L}_{CLAD+} = \mathcal{L}_{class}(x) + \mathcal{L}_{class}(x^+) + \lambda * \mathcal{L}_{con}(x, x^+, x^-) 
\label{eq:loss_clad+}
\end{equation}

Here, $\lambda$ is a hyperparameter for the weight that controls the importance of the contrastive term $\mathcal{L}_{con}$. Its magnitude controls the degree of background robustness learned by the model.

\subsection{Training}
\label{sec:training}

\begin{figure*}[ht]
\centering
\includegraphics[width=\textwidth]{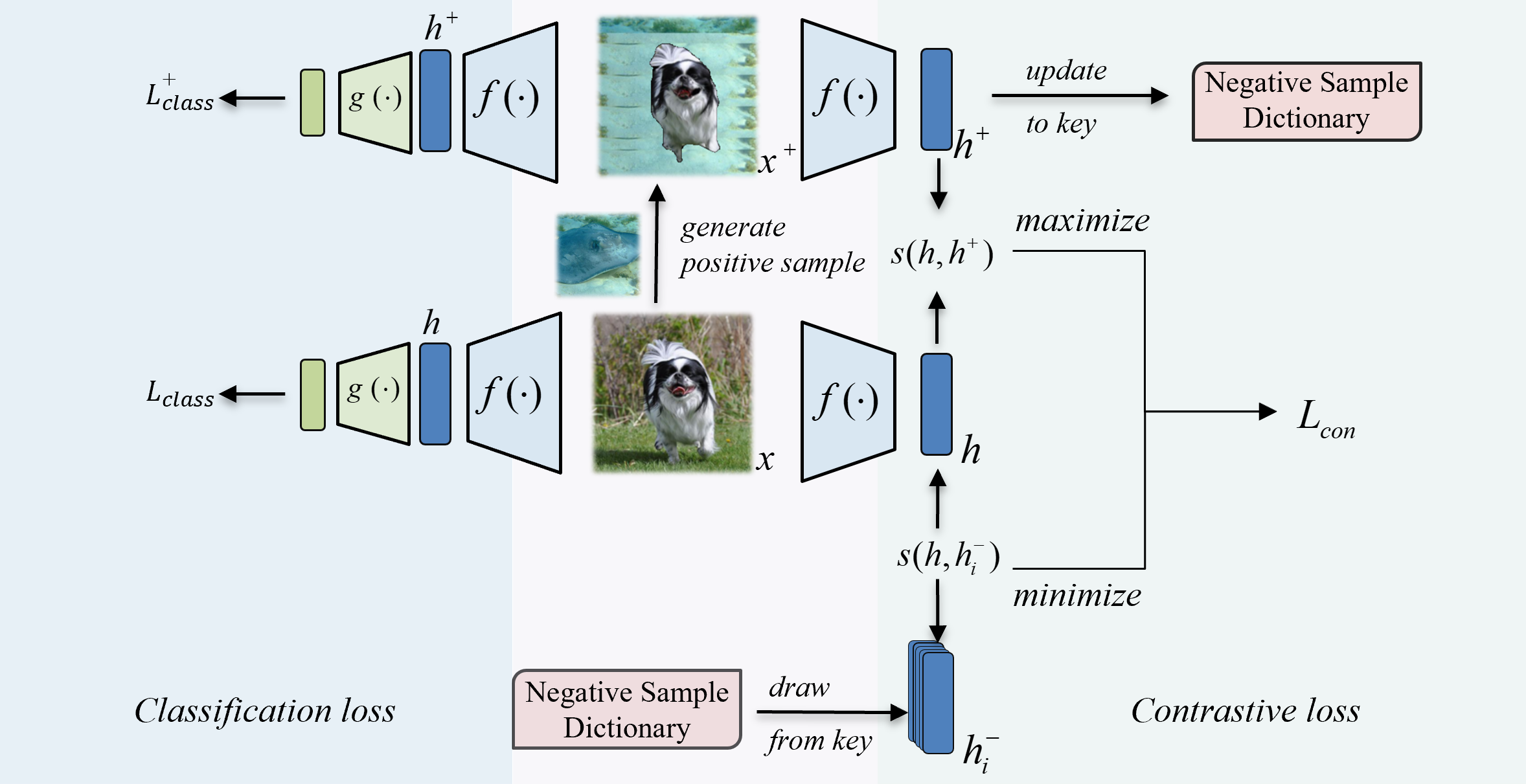}
\vspace{-1.5em}
\caption{Illustration of the proposed supervised learning with contrastive learning approach.}
\label{fig:flow_chart} 
\end{figure*}

As illustrated in Fig. \ref{fig:flow_chart}, for each batch, we generate positive samples. Then, the generated positive samples are used to update the negative sample dictionary. The classification loss is calculated for the anchor (and also for the positive sample for CLAD+). When calculating the contrastive loss, the negative samples are drawn accordingly from the negative sample dictionary based on the label of each anchor. The contrastive loss is finally calculated based on feature representations for the anchors, positive and negative samples.

\section{Experiment}
In this section, we present the results for CLAD and CLAD+ on the Background Challenge dataset \cite{xiao2020noise}. 

\subsection{Challenge Description}
Kai et al.\cite{xiao2020noise} initiated the Background Challenge \cite{xiao2020noise} dataset in 2020. The dataset aggregates a subset of images in ImageNet based on WordNet hierarchy \cite{miller1995wordnet} into 9 classes, creating the ImageNet-$9$ dataset. Several variations are made on the images' background or foreground in original ImageNet-$9$, as summarized in Table \ref{tab:dataset}.

\begin{table}[h!]
\centering
\resizebox{0.9\textwidth}{!} {
\begin{tabular}{@{}c|c|c|c@{}}
\toprule
Dataset & Foreground & Background & Summary \\ \midrule
\textsc{Original} & Original & Original & Original unaltered images \\
\textsc{Only-FG} & Original & None (Black) & Images with only the foreground (background removed) \\
\textsc{Only-BG-T} & None & Original & Images with only the background \\
\textsc{Mixed-Rand} & Original & Completely Random & Images with a random background \\
\textsc{Mixed-Same} & Original & Random-same-class & Images with a background from the same class \\ \bottomrule
\end{tabular}
}
\caption{Description of the variations of ImageNet-$9$\cite{xiao2020noise}}
\label{tab:dataset}
\end{table}

The goal of the Background Challenge is to achieve high accuracy on the \textsc{Mixed-Rand} dataset, where the background class is selected randomly and provides no information on image label. Intuitively, models with high background bias would suffer from low accuracy on this dataset. Additionally, the challenge also defines a metric to quantify the background bias: \textsc{BG-Gap}, which is defined as the accuracy gap between the \textsc{Mixed-Same} and \textsc{Mixed-Rand} datasets. The \textsc{BG-Gap} represents the performance drop due to background class signal change \cite{xiao2020noise}, or more intuitively, how much accuracy is actually gained by background bias. 
% The authors showed in \cite{xiao2020noise} that most state-of-the-art image classification models exhibit a poor generalizing ability to all of random background dataset variants with a significant BG-Gap. \\

\subsection{Experimental Settings}
\label{sec:experimental_details_background}
We adopt a ImageNet-pretrained ResNet-50 as our backbone \cite{xiao2020noise}. Adam \cite{kingma2014adam} is used as the optimizer with default settings ($\beta_1 = 0.9$ and $\beta_2 = 0.999$) and no weight decay is used. The total number of training epochs is 60, and the batch size is 64. The learning rate is set to be $1e^{-3}$ and decays to $1e^{-4}$ after 20 epochs. After trial and error, the hyperparameter $\lambda$ for the weight of the contrastive loss is set to $1$ (ablation in \ref{sec:analysis}) and the temperature parameter $\tau$ is set to 0.2. Data augmentations, including Random Resized Crop, Random Horizontal Flip, and Color Jitter, are used in our experiments. Note that for the generated positive samples, these conventional data augmentations are applied after the background augmentation. For each anchor, we construct one positive sample and draw 32 negative samples (details in supplementary material) from the negative sample dictionary.
In this section, we evaluate the performance of CLAD on the Background Challenge dataset. For comparison, we compare the performance of CLAD to  three baselines, which are trained in conventional, fully supervised settings, which include: 
\begin{itemize} 
\itemsep-0.5em 
    \item \texttt{Base(IN)} ImageNet-trained ResNet-50 with prediction mapped to ImageNet-$9$.
    \item \texttt{Base(IN9)} ResNet-50 trained on \textsc{Original} with a fully supervised setting.
    \item \texttt{Base(MR)} ResNet-50 trained on \textsc{Mixed-Rand} with a fully supervised setting.    
\end{itemize}

% In the following, \texttt{Base(IN)} and \texttt{Base(IN9)} will be referred to as in-distribution baselines. 
In addition we also compare to previous works on Background challenge dataset, the results of which are presented in Table \ref{table:acc_benchmark}.

\subsection{Accuracy}
CLAD and CLAD+ do not suffer any accuracy trade-off on the \textsc{Original} dataset compared to the baseline models (0.4\% and 0.1\% drop correspondingly). 
% \begin{wrapfigure}{r}{0.5\textwidth}
%     \centering
%     \includegraphics[width = 0.49\textwidth]{images/results.png}
%     \caption{Model performances on \textsc{Mixed-Same} (x-axis) and \textsc{Mixed-Rand} (y-axis) data. Models closer to the Identity dashed line has lower background bias.}
%     \label{fig:res}    
% \end{wrapfigure}
\begin{figure}
    \centering
    \includegraphics[scale=0.7]{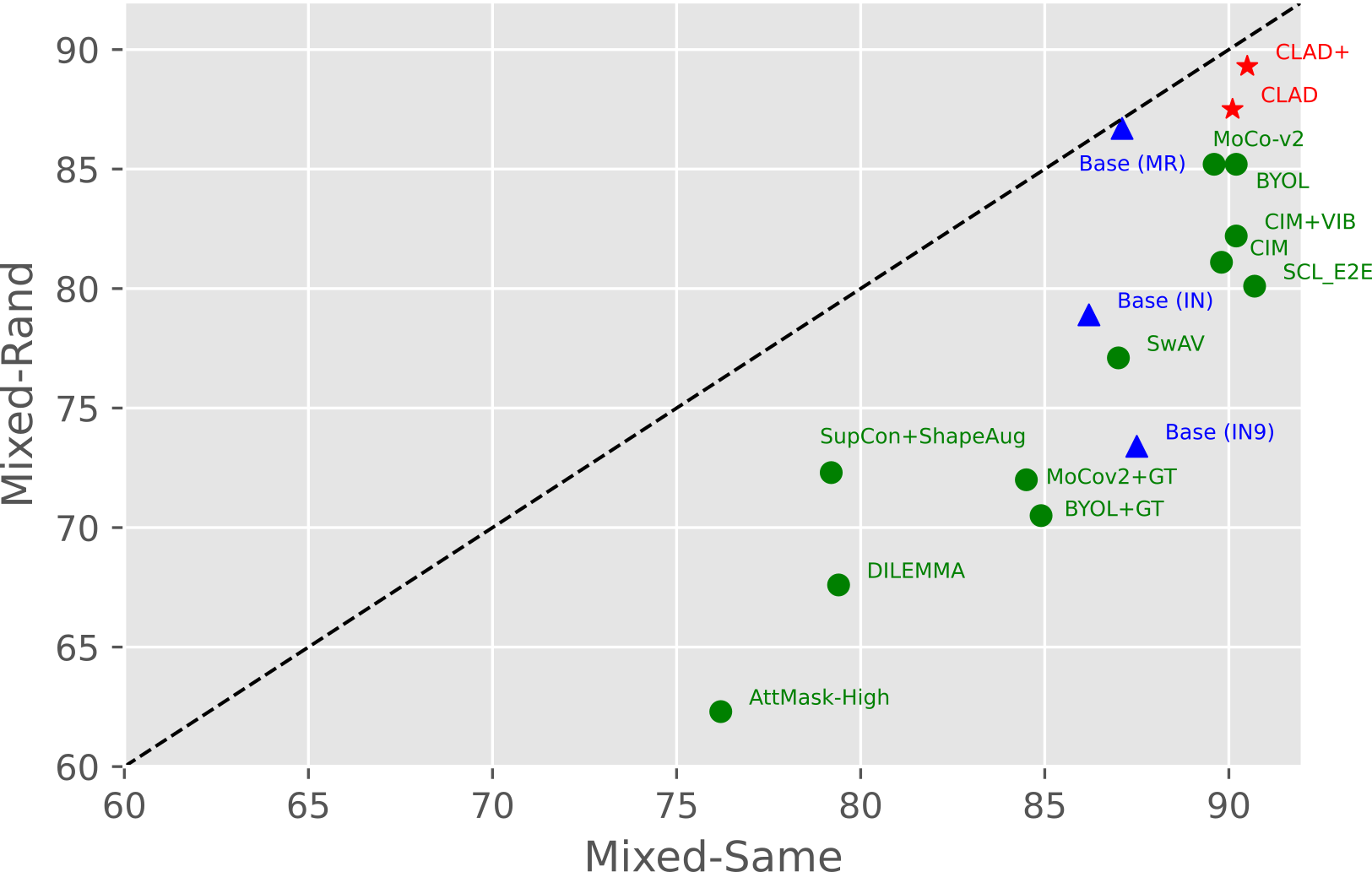}
    \caption{Model performances on \textsc{Mixed-Same} (x-axis) and \textsc{Mixed-Rand} (y-axis) data. Models closer to the Identity dashed line has lower background bias.}
    \label{fig:res}    
\end{figure}
Our method outperforms all previous benchmarks by a large margin (4.1\% for CLAD+ and 2.3\% for CLAD) on \textsc{Mixed-Rand} dataset, which is the most important indicator for the model's generalization ability to varying-background images. 

It is possible to have a very small \textsc{BG-GAP} as well as very low accuracy on both the \textsc{Mixed-Same} and \textsc{Mixed-Rand} datasets. However, that would not be reflective of the background bias or generalization ability of the model. Hence, we need high performance on both datasets along with a smaller gap between them, to have less background bias. We plot the accuracies of these datasets in Fig. \ref{fig:res}, wherein models that lie closer to the identity line have lower background bias. Additionally, the further right from the model's line, the higher its bias. We can see from the Figure that our models CLAD and CLAD+ have the best performance among all the models.

% Besides, we also observe significant better performance of our models on the \textsc{Only-FG} dataset with a margin of 6.3\% and 5.5\% respectively. 

% One thing to note is that the benchmarks in Table \ref{table:acc_benchmark} are trained on different datasets, so direct comparison on absolute accuracy might not be a fair comparison. However, \textsc{BG-Gap} is independent from the train set, making it an objective metric for model's background robustness. Again, our model has a much smaller \textsc{BG-Gap} compared with all the previous benchmarks (1.2\% and 2.6\% respective for our models, compared to previous state-of-art of 4.4\%). 

\begin{table*}[h!]
\centering
\resizebox{\textwidth}{!} {
\begin{tabular}{@{}llcccccc@{}}
\toprule
Type & \multicolumn{1}{c}{Model} & \textsc{Original}~$\uparrow$ & \textsc{Only-FG}~$\uparrow$ & \textsc{Mixed-Rand}~$\uparrow$ & \textsc{Mixed-Same}~$\uparrow$ & \textsc{Only-BG-T}~$\downarrow$ & \textsc{BG-Gap}~$\downarrow$ \\ \midrule
\multirow{3}{*}{Baselines} & Base (IN) \cite{xiao2020noise} & 96.2 & - & 76.3 & 82.3 & 17.8 & 6.0 \\
 & Base (IN9) & 96.0 & 86.0 & 73.4 & 87.5 & 42.9 & 14.1 \\
 & Base (MR) & 88.4 & 89.5 & 86.7 & 87.1 & 12.8 & 0.4 \\ \midrule
\multirow{11}{*}{Others} & CIM \cite{taghanaki2021robust} & 97.7 & - & 81.1 & 89.8 & - & 8.8 \\ 
 & SCL\textunderscore E2E \cite{taghanaki2021robust} & \textbf{98.2} & - & 80.1 & \textbf{90.7} & - & 10.6 \\
 & CIM+VIB \cite{taghanaki2021robust} & \textbf{97.9} & - & 82.2 & 90.2 & - & 8.0 \\
 & SupCon+ShapeAug\cite{lee2022improving} & - & - & 72.3 & 79.2 & - & 6.89 \\
 & MoCo-v2 (BG Swaps)\cite{ryali2021leveraging} & 95.2 & 87.5 & 85.2 & 89.6 & \textbf{11.4} & 4.4 \\
 & BYOL (BG Random)\cite{ryali2021leveraging} & 96.1 & 88.3 & 85.2 & 90.2 & 12.9 & 5.0 \\
 & SwAV (BG RM)\cite{ryali2021leveraging} & 95.3 & 86.8 & 77.1 & 87.0 & 18.2 & 9.9 \\
 & AttMask-High\cite{kakogeorgiou2022hide} & 89.8 & 75.2 & 62.3 & 76.2 & 15.3 & 9.9 \\
 & MoCov2+GT\cite{mo2021object} & 89.7 & 72,7 & 72.0 & 84.5 & 40.1 & 12.5 \\
 & BYOL+GT\cite{mo2021object} & 91.0 & 72.6 & 70.5 & 84.9 & 41.2 & 14.4 \\
 & DILEMMA\cite{sameni2022dilemma} & 91.8 & 77.8 & 67.6 & 79.4 & \textbf{9.3} & 10.2 \\ \midrule
\multirow{2}{*}{Ours} & CLAD+ & 95.6 & \textbf{94.6} & \textbf{89.3} & \textbf{90.5} & 22.6 & \textbf{1.2} \\
 & CLAD & 95.9 & \textbf{93.8} & \textbf{87.5} & 90.1 & 31.3 & \textbf{2.6} \\ \bottomrule
\end{tabular}
}
\caption{Accuracy ($\%$) comparison between CLAD, CLAD+ against baselines and benchmarks on the Background Challenge. `-' represents value missing in the references. Note that the models in this table are trained with different sizes of dataset and level of supervision, in this case the \textsc{BG-Gap} is the fairest comparison across all models indicating background bias.}
\label{table:acc_benchmark}
\end{table*}

\subsection{Analysis}
\label{sec:analysis}
\textbf{Feature Consistency}
We estimate the percentage of encoded foreground information by calculating the features' cosine similarity between image pairs sharing the same foreground. This metric can also be intuitively reflect as how much of the features are extracted from the foreground. We also define a more direct metric, decision consistency, which summarizes the fraction of consistent decisions after background change. This can be expressed as $ \frac{1}{N}\sum_{i=1}^{N} \mathbbm{1}(\argmax g(x_i) = \argmax g(\hat{x}_i)) $, where, $g(.)$ is the classifier, $(x_i, \hat{x}_i)$ represent image pairs with same foreground but different background. The higher the decision consistency, the smaller the effect of background changes on the models' decisions. For details, see Table \ref{table:cosine_similarity_background}. 

% \textit{Further, we note that the decision consistency of CLAD and CLAD+ between \textsc{Only-FG} and \textsc{Mixed-Rand} datasets are both higher than their pure accuracy in the \textsc{Mixed-Rand} dataset, meaning that some of their errors on \textsc{Mixed-Rand} is actually not due to background influence. As a comparison, the decision consistency metric for \texttt{BL(MR)} is lower than the pure accuracy in the \textsc{Mixed-Rand} dataset.

% As we can see, \texttt{Base(IN9)} extracts very different features for the Original dataset and the other three datasets, with the averaged feature similarity lower than $70\%$. This explains its poor accuracy and decision consistency (just above 80\%) with changing background: a large proportion of its features are extracted from image background, whose spurious nature limits its generalization ability under changing backgrounds.}

\begin{table}[h]
\centering
\resizebox{0.5\textwidth}{!} {
\begin{tabular}{lcc}
\hline
\multicolumn{1}{c}{Model} & Feature Similarity & Decision Consistency \\ \hline
Base (IN9)                  & 0.795              & 0.800                \\
Base (MR)                   & 0.864              & 0.864                \\ \hline
CLAD+                    & \textbf{0.920}              & \textbf{0.969}                \\
CLAD                      & 0.914              & 0.915                \\ \hline 
\end{tabular}}
\caption{Feature similarity and decision consistency between \textsc{Original} and \textsc{Mixed-Rand} datasets. CLAD and CLAD+ can extract features over 90\% similar to the extracted features before background variation, meaning that a large amount of the features they learned from the original images are from the foreground, well explaining their performance on the background challenge dataset.}
\label{table:cosine_similarity_background}
\end{table}

\begin{figure*}[h]
\centering
\includegraphics[width=0.80\textwidth]{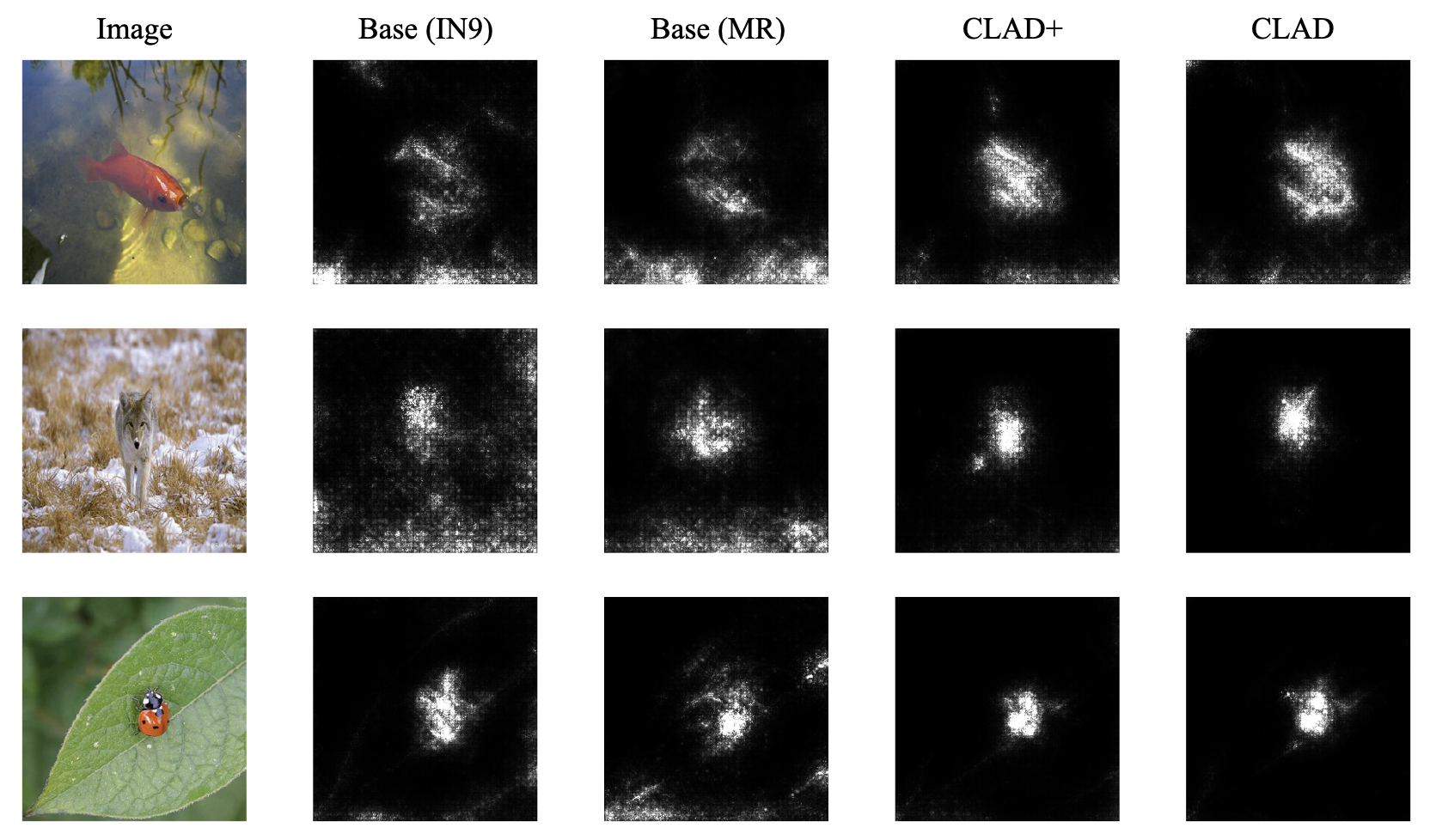}
\vspace{-1.5em}
\caption{Saliency maps of CLAD+ and CLAD, compared with two baseline models.}
\label{fig:saliency}
\end{figure*}

\textbf{Interpretability: } Saliency map provides intuitive illustration for models' areas of focus in images. Fig.~\ref{fig:saliency} illustrates the SmoothGrad\cite{smilkov2017smoothgrad} saliency maps of the CLAD and CLAD+, compared with two baseline models. It shows that the saliency maps for CLAD+ and CLAD focus more on the foreground object with a much cleaner saliency map than \texttt{Base(IN9)} and even \texttt{Base(MR)}. An interesting observation is that, the saliency map of \texttt{Base(IN9)} and \texttt{Base(MR)} on the \texttt{wolf} image (second row) shows these baseline models rely on background snow for identifying wolf, which is a well-known example for CNN's background bias \cite{szegedy2015going}. CLAD+ and CLAD are able to identify \texttt{wolf} while ignoring background snow, relatively better than base models.

% \pascal{could it be interesting to measure robustness too? to CC or adv perturbations? That would respond to one of the motivations in the introduction, that is to remove bias and make models more robust (or leave less 'space' for perturbations.}

% \pascal{would be interesting to extend the method to texture, or other biases (not all of them though) - to show that the method is general (minimal description of the necessary changes to the model explained earlier for background bias, would be necessary then - it can be shortly added here, in the corresponding experiment subsection.}

\begin{figure*}[h]
\centering
\includegraphics[width=\textwidth]{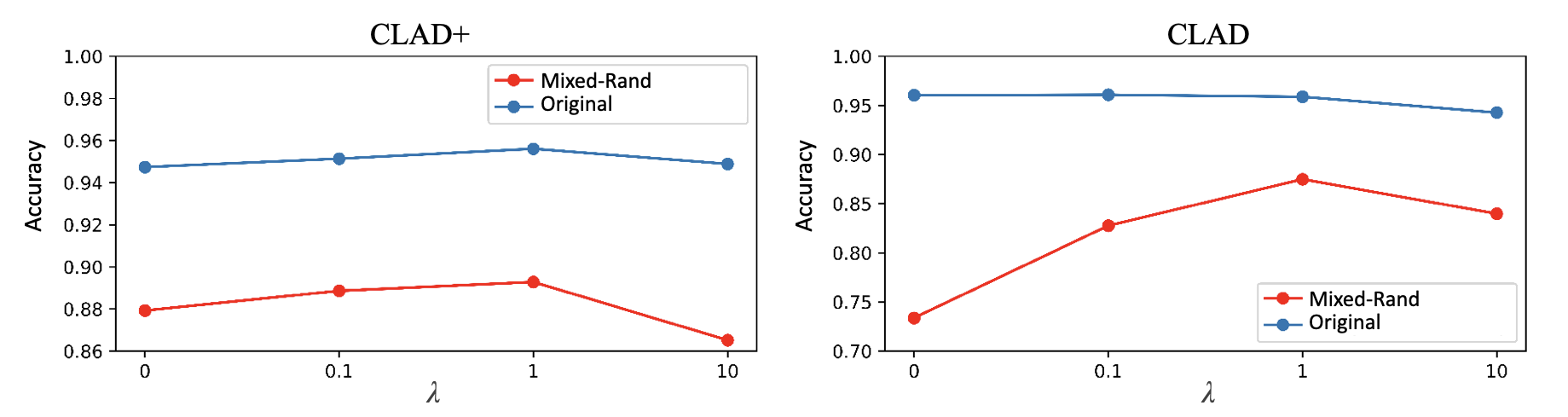}
\vspace{-1.5em}
\caption{Accuracy on \textsc{Mixed-Rand} dataset and Original \textsc{ImageNet-$9$} with respect to different values of $\lambda$}
\label{fig:ablation_lambda} 
\end{figure*}

\textbf{Importance of Contrastive Loss:} The hyper-parameter $\lambda$ is the weight of the contrastive loss term in our overall loss function. In Fig. \ref{fig:ablation_lambda} we show that the magnitude of $\lambda$ determines the background robustness of the CLAD model. This Figure presents the varying accuracy on \textsc{Mixed-Rand} dataset, as well as \textsc{Original ImageNet-$9$}, with increasing $\lambda$. The CLAD models do not have any performance deterioration when the contrastive loss is introduced with equal weight as the classification loss. Its performance on \textsc{Original} dataset remains the same while increasing on \textsc{Mixed-Rand}. However, if the value of $\lambda$ is further increased, i.e., the contrastive loss becomes more important than the supervised losses ($\lambda > 1$), then there is a performance deterioration. This indicates that we need a balanced mix of both supervised and contrastive losses for ideal performance. 

% While this trend is more obvious for CLAD than CLAD+, we observe that an intermediate value of $\lambda$ (around 1) gives optimal performance on \textsc{Mixed-Rand} dataset. This implies that contrastive loss is necessary for good performance. Another key observation is that, 
% While the accuracy on clean \textsc{ImageNet-9} of CLAD stays stable for a $\lambda$ value lower than 10, this accuracy even increased for CLAD+.

% \subsection{Negative samples}

% \subsection{Robustness on Common Corruptions}
% We test the robustness of CLAD and CLAD+ on 

\section{Conclusion}
\label{sec:conclusion}
Through our work, we present a novel contrastive learning-based approach for background debiasing called CLAD. It samples background-debiased contrastive pairs efficiently. Our work showcases state-of-the-art performance on the Background Challenge dataset. We also show an analysis of our model's features, which explain its superior performance compared to the standard trained model. Further, we empirically demonstrate the need for proper balance between contrastive and supervised losses for the effective debiasing of the model. As a result, training with the proposed contrastive learning method reduces the importance of image background and texture information in the decision-making process of CNN models. Theoretically, this approach works for any discriminative feature pairs, and we took \emph{foreground} vs. \emph{background} and \emph{shape} vs. \emph{texture} (in supplementary material) as an example. In future works, we could further investigate how to extend this approach to other pairs of discriminative features and hopefully guide the CNNs to make decisions based on similar features as humans, thereby improving generalization ability.

\bibliography{main}
% \end{document}

\clearpage

% \title{Supplementary}
% \maketitle

\setcounter{section}{0}
\renewcommand{\thesection}{\Roman{section}}

\begin{center}
    \section*{Supplementary}
\end{center}

\section{Effect of Negative Sample Dictionary}

\subsection{Ablation for Negative Sample Dictionary}

To validate the effectiveness of negative sample dictionary, we conduct an ablation study. We originally create negative samples, which contain background associated with the anchor's foreground class. However, for this ablation, we create trivial negative samples, which contain background from random classes and are not necessarily matched with anchor's foreground class. These trivial negative samples are shared across all anchors. 

% here we use trivial negative samples for calculating the contrastive loss instead of using the samples with similar background information. 
% All the experiment settings remain the same, except that the negative samples are stored uniformly inside a single queue.
Table \ref{table:acc_trivial_neg} presents the performance comparison between CLAD and CLAD+ models, with their respective counterparts where trivial negative samples are used, denoted as CLAD (Trivial) and CLAD+ (Trivial). 

\begin{table*}[h!]
\centering
\resizebox{\textwidth}{!} {
\begin{tabular}{lcccccc}
\hline
\multicolumn{1}{c}{Model} & \textsc{Original}$\uparrow$ & \textsc{Only-FG}$\uparrow$        & \textsc{Mixed-Rand}$\uparrow$     & \textsc{Mixed-Same}$\uparrow$ & \textsc{Only-BG-T}$\downarrow$ & \textsc{BG-Gap}$\downarrow$      \\ 
\hline
CLAD                      & \textbf{95.9}                     & \textbf{93.8}      & \textbf{87.5}     & \textbf{90.1}                       & \textbf{31.3}                        & \textbf{2.6} \\
CLAD (Trivial)            & 95.5                     & 93.0                           & 85.3                           & 88.9                       & 37.2                        & 3.6                           \\
\hline
CLAD+                     & \textbf{95.6}                     & 94.6     & \textbf{89.3}      & \textbf{90.5}                      & \textbf{22.6}                        & \textbf{1.2} \\
CLAD+ (Trivial)           & 95.4                     & \textbf{94.7}                           & 89.1                           & 90.3                       & 24.7                        & \textbf{1.2}                           \\ 
\hline
\end{tabular}
}
\caption{Accuracy ($\%$) comparison between CLAD, CLAD+ against their counterparts where trivial negative samples are used.}
\label{table:acc_trivial_neg}
\end{table*}

We observe that both CLAD and CLAD+ indeed perform better than their counterparts which use trivial negative samples.

\subsection{Effect of Different Number of Negative Samples}

\begin{figure*}[h]
\centering
\includegraphics[width=0.9\textwidth]{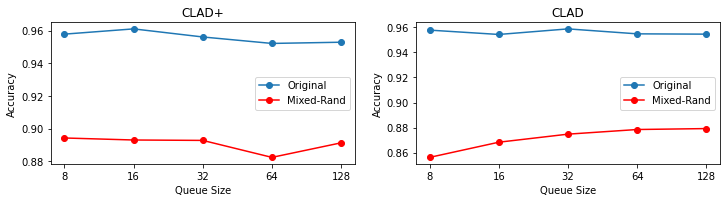}
\vspace{-1.5em}
\caption{Ablation for the size of the queue in negative sample dictionary. Small queue sizes suffer from insufficient negative samples while large queue size suffers from deterioration of \textsc{Original} accuracy and increased computational costs.}
% The larger the queue the more time they stay in it and hence less 'fresh' neg. samples, which is bad because the latent representation is stored and that is dependent on the state of the network ... }
\label{fig:queue_ablation}
\end{figure*}

In Fig. \ref{fig:queue_ablation}, we show the accuracy of using different queue sizes (which is also the number of negative samples) for the dictionary. We choose the size to be 32 as it is the best trade-off for both models.

\section{Foreground Segmentation and Scalability}

In our experiments, following the Background Challenge dataset, we used 10 iterations of GrabCut \cite{rother2004grabcut} to segment the images' foreground using bounding box(bb) information. Instead of Grabcut which relies on bb, we tested using pre-trained U2-Net \cite{qin2020u2} to segment the foreground (Results in Table \ref{table:fg_method}). We can see that the accuracy gap between the two methods on the \textsc{Original} and \textsc{Mixed-Rand} dataset is within 1\%, and results with U2-Net still beat all previous benchmarks on the Background Challenge. Since the performance does not change significantly, we can replace GrabCut with scalable methods like U2-Net, hence improving the scalability of our method.

In previous works in Table 2 of the main paper, \cite{lee2022improving, ryali2021leveraging} also use foreground segmentation supervision, making them fair comparisons to ours.

\begin{table}[h]
\centering
\tiny{
\resizebox{0.7\textwidth}{!}{
\begin{tabular}{lccccccc}
\hline
Model                  & \multicolumn{1}{l}{FG segmentation} & \multicolumn{1}{l}{Original~$\uparrow$} & \textsc{Mixed-Rand}~$\uparrow$ \\ \hline
Base (IN9)             & \      -       & \textbf{96.0}    & 73.4  \\ \hline
\multirow{2}{*}{CLAD+} & GrabCut       & 95.6    & \textbf{89.3}  \\
                       & U2-Net         & \textbf{96.0}    & 88.3  \\ \hline
\multirow{2}{*}{CLAD}  & GrabCut       & 95.9    & 87.5  \\
                       & U2-Net         & 95.8    & 87.1  \\ \hline
\end{tabular}}
\caption{Accuracy ($\%$) comparison for using GrabCut and U2-Net as foreground segmentation methods.}
\label{table:fg_method}
}
\end{table}

\section{Potential Foreground Positional Bias} 
The Background Challenge dataset may have centered foreground bias. Therefore, we use \texttt{FiveCrop} of \textsc{PyTorch} to crop from the corners of the ImageNet-9 dataset and create foreground shift from the center. We report the averaged accuracy drop (Table \ref{table:crop}). Our methods do not suffer from positional bias compared with baseline models. Contrastive learning penalizes positional shift bias because it enforces similarity between randomly augmented (main paper Sec.4.2) positive pairs.

\begin{table}[h!]
\centering
\resizebox{0.7\textwidth}{!}{
\begin{tabular}{lcccc}
\hline
\multicolumn{1}{c}{} & Base (IN9) & Base (MR) & CLAD+ & CLAD  \\ \hline
Accuracy drop ($\%$)~$\downarrow$            & 4.6      & 7.9     & \textbf{4.1} & 4.5 \\ \hline
\end{tabular}}
\caption{Averaged accuracy drop ($\%$) after corner cropping on Original dataset.}
\label{table:crop}
\end{table}

\section{Mitigate Texture Bias}

% Could we explain it like: overfitting to the background is mostly due to overfitting to the background texture, which would explain why the shape-biased model also have less background bias
We show in this section how our method can be extended to texture biases. Previous works have shown that CNNs are biased towards local texture, instead of global shape \cite{geirhos2018imagenet, baker2018deep, brendel2019approximating, hermann2020origins}. CNN's over-reliance on texture limits both its connection to human vision systems and its vulnerability to OOD data with texture-shape cue conflict \cite{hermann2020origins}. Increasing CNN's shape bias would improve CNN's robustness towards a wide range of image distortions \cite{geirhos2018imagenet}. 

In this part, we show that the CLAD approach can be extended to other discriminative features. As an example, we show that it successfully reduces CNN's texture bias.

\begin{figure*}[h]
\centering
\includegraphics[width=\textwidth]{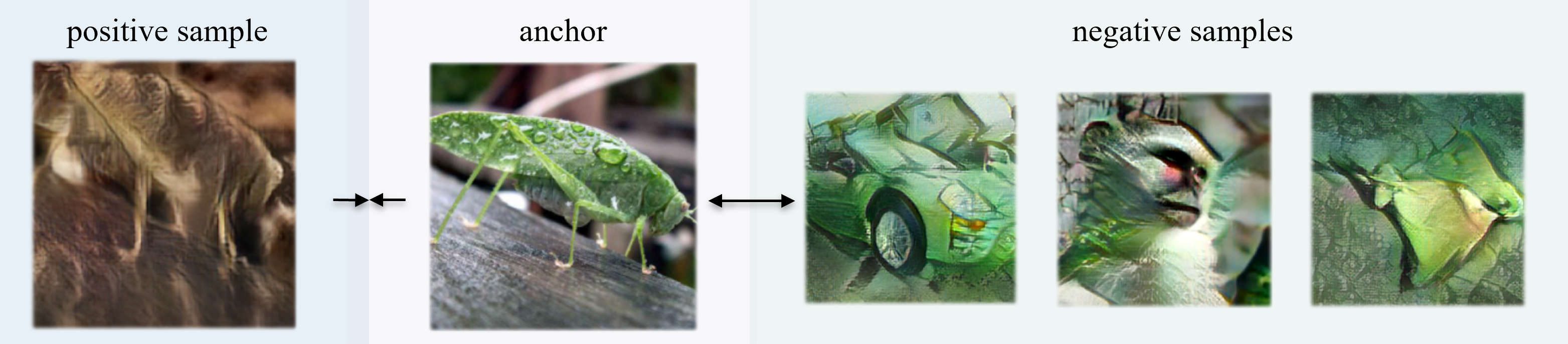}
\vspace{-1.2em}
\caption{Example for sampling contrastive pairs for reducing texture bias}
\label{fig:shape_pairs} 
\end{figure*}

For the training scheme, we adopt the same approach as before and again experiment on the ImageNet-$9$ dataset, with the exception of how we generate contrastive pairs. We follow the basic idea that undesired discriminative feature (texture in this case) should be shared between negative sample pairs, while desired discriminative feature (shape) should be shared between positive sample pairs. In practice, when generating the cue-conflict images, we use the AdaIN \cite{huang2017arbitrary} algorithm to modify the anchor's texture information. An example of the contrastive pairs used in our model (S-CLAD, S-CLAD+) for reducing texture bias is shown in Fig. \ref{fig:shape_pairs}. 

\vspace{0.5em}

\textbf{Datasets:} We evaluate models' shape bias on two datasets: \textsc{Stylized} ImageNet-$9$ and ImageNet-$9$-\textsc{Sketch}.  We generate the \textsc{Stylized} ImageNet-$9$ using the same algorithm (AdaIN \cite{huang2017arbitrary}) as the original Stylized ImageNet \cite{geirhos2018imagenet}. ImageNet-$9$-\textsc{Sketch} is created by mapping the classes in ImageNet-Sketch \cite{wang2019learning} to classes in ImageNet-$9$. Models with high shape bias are expected to have better accuracy on these datasets, as the texture information in these datasets is either randomized or removed and hence provides no useful information on the class label. 

The performance of S-CLAD+ and S-CLAD is compared against two baselines, \texttt{Base(IN9)} and \texttt{Base(SIN9)}, where the latter is a baseline model trained on stylized images from ImageNet-$9$ in a fully supervised setting.
The results are presented in Table \ref{table:shape_acc}. We can see that, both S-CLAD and S-CLAD+ outperform the \texttt{Base(IN9)} baseline with a large margin on \textsc{Stylized} ImageNet-$9$ and ImageNet-$9$-\textsc{Sketch}, indicating their texture bias is mitigated. Moreover, we again observe that there is almost no accuracy trade-off on the \textsc{Original} ImageNet-9 for S-CLAD and S-CLAD+, whereas \texttt{Base(SIN9)} suffers from performance drop on the \textsc{Original} ImageNet-9. Note that, no sketch images are included in S-CLAD+ and S-CLAD's training process, and they still have a performance gain of around 20\% on the ImageNet-$9$-\textsc{Sketch} dataset. This performance gain is because the model is focusing more on shape information.
\begin{table}[h]
\centering
\resizebox{0.8\textwidth}{!} {
\begin{tabular}{lccc}
\toprule
\multicolumn{1}{c}{Model} & \textsc{Original} ImageNet-$9$ & \textsc{Stylized} ImageNet-$9$ & ImageNet-$9$-\textsc{Sketch} \\ \toprule
Base (IN9)                  & \textbf{96.0}    & 53.6              & 40.1              \\ 
Base (SIN9)                  & 91.5            & 75.1              & 58.0             \\ \hline
S-CLAD                      & 95.1              & 74.4              & \textbf{61.0}              \\ 
S-CLAD+                     & 95.5              & \textbf{76.7}    & 58.6              \\ \bottomrule
\end{tabular}}
\caption{Accuracy comparison for S-CLAD+ and S-CLAD, against \texttt{Base(IN9)} on three datasets. The accuracy on \textsc{Stylized} and \textsc{Sketch} dataset indicates model's shape bias.}
\label{table:shape_acc}
\end{table}

\section{Cross-Evaluation}
In this section, we cross-evalutate whether background-debiased models (CLAD and CLAD+) generalize better to the texture variation, and vise-versa. Firstly, we evaluate the shape bias of CLAD and CLAD+, compared against \texttt{Base(IN9)} on \textsc{Stylized} ImageNet-$9$ and ImageNet-$9$-\textsc{Sketch}.

\begin{table}[h]
\centering
\resizebox{0.8\textwidth}{!} {
\begin{tabular}{lccc}
\toprule
\multicolumn{1}{l}{Model} & \textsc{Original} ImageNet-$9$ & \textsc{Stylized} ImageNet-$9$ & ImageNet-$9$-\textsc{Sketch} \\ \toprule
Base (IN9)                  & \textbf{96.0}    & 53.6              & 40.1              \\ \hline
CLAD                      & 95.9              & \textbf{54.3}              & 39.7              \\ 
CLAD+                     & 95.6              & 53.7              & \textbf{41.2}              \\ \bottomrule
\end{tabular}}
\caption{Evaluation of shape bias for CLAD, CLAD+, compared against \texttt{Base(IN9)}.}
\label{table:bg_shape_acc}
\end{table}

The results show a minor improvement by CLAD and CLAD+ on \textsc{Stylized} ImageNet-$9$ and ImageNet-$9$-\textsc{Sketch} respectively. 
This can explained due to the fact that by removing background bias, we focus more on both the foreground shape and texture information. Hence, our model may still use the texture information from the foreground for classification along with its shape. Thus, its performance on datasets which transform the texture of the foreground and background together is not improved drastically. 
This also implies that background debiasing alone is not sufficient for texture debiasing. 

\begin{table*}[h!]
\centering
\resizebox{\textwidth}{!} {
\begin{tabular}{llccccc}
\hline
\multicolumn{1}{c}{Model} & \textsc{Original}$\uparrow$ & \textsc{Only-FG}$\uparrow$        & \textsc{Mixed-Rand}$\uparrow$     & \textsc{Mixed-Same}$\uparrow$ & \textsc{Only-BG-T}$\downarrow$ & \textsc{BG-Gap}$\downarrow$      \\ 
\hline
Base (IN9) & \textbf{96.0} & 86.0 & 73.4 & 87.5 & 42.9 & 14.1 \\ \hline
S-CLAD       & 95.0 & \textbf{87.5} & \textbf{78.9} & \textbf{89.3} & 40.9 & \textbf{10.4} \\
S-CLAD+      & 95.5 & 87.4 & 78.1 & 88.5 & \textbf{38.1} & \textbf{10.4}                           \\
\hline
\end{tabular}
}
\caption{Evaluation of S-CLAD, S-CLAD+ on Background Challenge, compared against \texttt{Base(IN9)}.}
\label{table:acc_shape_on_bg}
\end{table*}

We then test S-CLAD and S-CLAD+ on the Background Challenge datasets \cite{xiao2020noise}. As shown in Table. \ref{table:acc_shape_on_bg}, we find that inducing shape bias helps mitigate background bias to some extent. This can be explained by the increased focus on shape of the object which is usually in the foreground, while also ignoring the background information. 
However, texture-debiased model, S-CLAD and S-CLAD+ alone are not sufficient to reproduce our state of the art results we had from CLAD and CLAD+ on Background Challenge.

\end{document}